\documentclass[11pt]{article}

\usepackage{hyperref}
\usepackage{setspace}
\usepackage{palatino}
\usepackage{graphicx}
\usepackage{float}
\usepackage{titling} % drop vertical space before the title

\usepackage{lscape}
\usepackage{amsmath}
\usepackage{amssymb}

\fontfamily{ppl}\selectfont 

\usepackage[american]{babel}

\usepackage[natbibapa]{apacite} % to enable '\citet' and '\citep' macros

\usepackage{breakcites}
\usepackage{graphics}
\usepackage{multirow}
\usepackage{booktabs}
\usepackage{caption}
\usepackage{subcaption}

\usepackage{tablefootnote}

\usepackage{tablefootnote}

\usepackage{amsthm}

\theoremstyle{definition}
\newtheorem{definition}{Definition}[section]

\theoremstyle{remark}

\usepackage{flafter} % prevents figures ever floating backwards up the current page

\usepackage{lmodern}

\usepackage[utf8]{inputenc}

\usepackage[switch]{lineno}

\title{Detecting Ongoing Events Using Contextual Word and Sentence Embeddings
}

\author{Mariano Maisonnave\\% Name author
    {Universidad Nacional del Sur} \\
    \href{mailto:mariano.maisonnave@cs.uns.edu.ar}{\texttt{mariano.maisonnave@cs.uns.edu.ar}} %% Email author 1 
\and Fernando Delbianco\\% Name author
    {Universidad Nacional del Sur} \\
    \href{mailto:fernando.delbianco@uns.edu.ar}{\texttt{fernando.delbianco@uns.edu.ar}} %% Email author 2
\and Fernando Tohmé\\% Name author
    {Universidad Nacional del Sur} \\
    \href{mailto:ftohme@criba.edu.ar}{\texttt{ftohme@criba.edu.ar}}%% Email author 3
\and Ana Maguitman\\% Name author
    {Universidad Nacional del Sur} \\
    \href{mailto:agm@cs.uns.edu.ar}{\texttt{agm@cs.uns.edu.ar}}%% Email author 3
\and Evangelos Milios\\% Name author
    {Dalhousie University} \\
    \href{mailto:eem@cs.dal.ca}{\texttt{eem@cs.dal.ca}}%% Email author 3
    }
    
\date{}

\begin{document}

% %%%%%%%%%%%%%%%%%%%%%%%%%%%%%%%%%%%%%%%%%%%%%%%%%%%%%%%%%%
% %%%%%%%%%%%%%%%%%%%%%%%%%%%%%%%%%%%%%%%%%%%%%%%%%%%%%%%%%%
% ABSTRACT
% %%%%%%%%%%%%%%%%%%%%%%%%%%%%%%%%%%%%%%%%%%%%%%%%%%%%%%%%%%
% %%%%%%%%%%%%%%%%%%%%%%%%%%%%%%%%%%%%%%%%%%%%%%%%%%%%%%%%%%
{\setstretch{.8}
\maketitle
% %%%%%%%%%%%%%%%%%%
\begin{abstract}

This paper introduces the Ongoing Event Detection (OED) task, which is a specific Event Detection task where the goal is to detect ongoing event mentions only, as opposed to historical, future, hypothetical, or other forms or events that are neither fresh nor current. Any application that needs to extract structured information about ongoing events from unstructured texts can take advantage of an OED system. The main contribution of this paper are the following:  (1) it introduces the OED task along with a dataset manually labeled for the task; (2) it presents the design and implementation of an RNN model for the task that uses BERT embeddings to define contextual word and contextual sentence embeddings as attributes, which to the best of our knowledge were never used before for detecting ongoing events in news; (3) it presents an extensive empirical evaluation that includes (i) the exploration of different architectures and hyperparameters, (ii) an ablation test to study the impact of each attribute, and (iii) a comparison with a replication of a state-of-the-art model. The results offer several insights into the importance of contextual embeddings and indicate that the proposed approach is effective in the OED task, outperforming the baseline models.

% END CONTENT ABS------------------------------------------

\noindent

\end{abstract}
}

\section{Introduction}
\label{sec:introduction}
  
The Information Extraction (IE) task consists in extracting structured information from unstructured natural language texts. Event Extraction (EE) is a subtask of IE, in which the goal is to detect and retrieve real-world events from those texts. An EE system usually performs two different steps to complete the extraction of the events. The first step is to identify the event trigger, which is the word that most clearly expresses the occurrence of an event, and classify it into one of the predefined event types. This step is called Event Detection (ED).  The second step is to extract the arguments of the events. The ED task has been addressed in the literature, both as a standalone task \citep{nguyen2018graph,liu2018event,duan2017exploiting,nguyen2016modeling,nguyen2015event, nguyen2016two,wu2014zero,jagannatha2016bidirectional} and as a part of an EE system \citep{zhang2019joint,liu2018jointly,sha2018jointly,huang2017zero,nguyen2016joint,chen2015event,li2013joint,ahn2006stages}. 
There is a great incentive to study ED systems not only because of their direct use in several applications but also because any improvement in an ED system will impact directly on the performance of any EE system implemented with it.
EE and ED systems are crucial for any application or domain that needs structured information and relies on a large corpus of unstructured data. Some examples of this are question-answering systems~\citep{sha2018jointly} and text summarization systems~\citep{LEE2003431}. These systems are also useful for generating reports of the information available for a domain~\citep{ADEDOYINOLOWE2016351}. These reports can help an expert to make decisions or create policies to address an issue.

% >>
In this paper, we define and address the OED task as part of a broader project that aims at detecting ongoing real-world events and other relevant variables from news and social media with the ultimate goal of learning causal models~\citep{maisonnave2020assessing}. 
In the broader project, we use time series-related techniques from Econometrics to learn such causal models~\citep{granger1969investigating}. 
To build those time series, we require (i) all the event mentions in the news and (ii) the time those events took place. These two requirements defined the needs of our event detection task. Because of (i), we require only the detection of event mentions (trigger detection), not the event arguments nor the participants. Because of (ii), we require the event to be ongoing when reported (discarding past, future and hypothetical events). If the event is ongoing at the moment it is being reported, the news article's date can be used as the moment of occurrence of the event. These two requirements result in the definition of a proposed novel task, to which we refer to as the Ongoing Event Detection (OED) task. 

\begin{figure}[!ht]
\begin{subfigure}{.48\textwidth}
  \centering
  % include first image
  \includegraphics[width=1\linewidth]{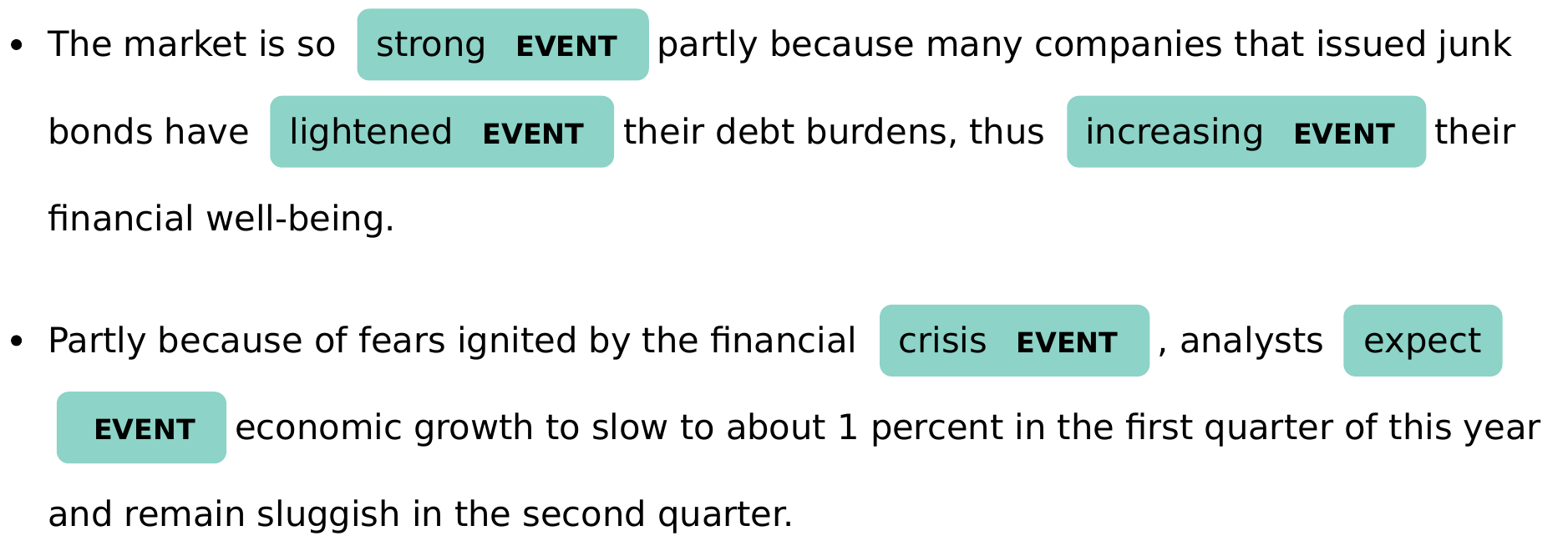}  
  \caption{}
  \label{fig:usecase_diagrama}
\end{subfigure}
\begin{subfigure}{.48\textwidth}
  \centering
  \includegraphics[width=0.8\linewidth]{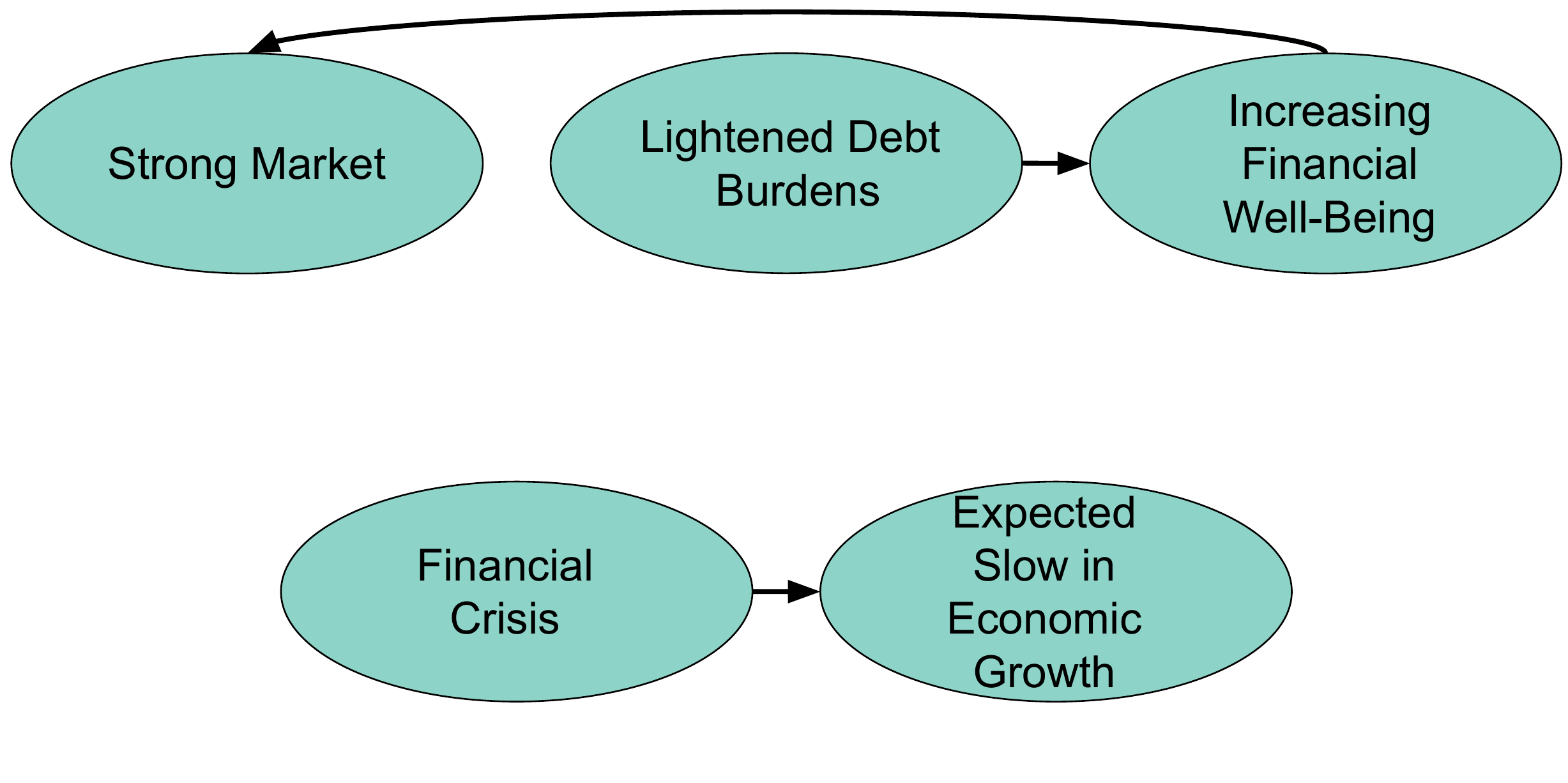} 
  \caption{}
  \label{fig:usecase_diagramb}
\end{subfigure}

    \caption{Two use cases for the OED tool, applied to two sentences (a). A causal graph manually extracted from the examples (b). }
    \label{fig:usecase_diagram}
\end{figure}

We divided the broader project into two steps, (1) the OED task  and (2) the time series construction and causal structure learning from those time series.
 The two steps involved in building the tool are depicted in Figure \ref{fig:usecase_diagram}. The work reported in this paper focuses on the first step only, i.e., on the OED task.

Motivated by the limitations of previous proposals to address our specific requirements, we outlined a definition of event (described in Section \ref{sec:OEDTask}) that, on the one hand, is not limited by a fixed taxonomy of events and, on the other hand, is centered on ongoing events only. Guided by this definition, we manually labeled a dataset for training and testing purposes (described in Section \ref{sec:dataset}). Using this dataset, we developed an RNN model for event prediction (described in Section \ref{sec:rnn}). Since there are no previous studies on this dataset, we also implemented two baselines for comparison. First, we implemented a simple OED model based on a classical approach (SVM model). Second, we replicated a baseline (CNN model) from the state-of-the-art in ED  to apply to our task~\citep{nguyen2015event}. Both baselines are described in Section \ref{sec:baseline}. The code for the proposed and baseline models, as well as the dataset, are made available to the research community for reproducibility and data reuse.\footnote{The code is available at \href{https://cs.uns.edu.ar/~mmaisonnave/resources/ED\_code/}{https://cs.uns.edu.ar/$\sim$mmaisonnave/resources/ED\_code/}, and the dataset is available at \href{https://cs.uns.edu.ar/~mmaisonnave/resources/ED\_data/}{https://cs.uns.edu.ar/$\sim$mmaisonnave/resources/ED\_data/}}. Results, discussion and conclusions are presented at the end of the paper in Sections \ref{sec:experiments}, \ref{sec:discussion}, \ref{sec:conclusions}, respectively.

The contributions of this paper  can be summarized as follows. 
\begin{enumerate}
    \item
    Firstly, we define the OED task for the first time, and we present a manually labeled dataset for the task. The task proved to be a promising direction for implementing the first step of the framework for causal structure learning depicted in Figure \ref{fig:usecase_diagram}. 
    \item 
    Secondly, we design and implement an RNN model for the OED task that includes BERT embeddings as features. The use of BERT embeddings for the task is extensively studied in this work, and the results demonstrate the usefulness of context-sensitive embeddings for this task. We also elaborate on a baseline by replicating a state-of-the-art model \citep{nguyen2015event} on our data, as well as a classical SVM model. The proposed RNN model outperforms the evaluated baselines. 
    \item 
    Finally, we present an extensive empirical evaluation with different  architectures and hyperparameters for both the baseline and our model. We also perform an ablation test to measure the impact of each feature on the performance. The full code of the experiments is made available to allow reproducibility.

\end{enumerate}

\section{Background and Related Work} \label{sec:background}

The most widely used dataset for EE is the ACE 2005 Multilingual Training Corpus~\citep{walker2006ace}, which contains the complete set of English, Arabic, and Chinese training data for the 2005 Automatic Content Extraction (ACE) technology evaluation~\citep{doddington2004automatic}.  There are other EE and ED datasets available, such as TimeML~\citep{pustejovsky2003timeml} and SentiFM~\citep{jacobs2018economic}. However, to the best of our knowledge, all these datasets have a fixed taxonomy of valid event types. 

Three main disadvantages arise when using existing EE datasets from the literature. First, as already mentioned, these datasets limit the number of events to a fixed set of possible events (those available in the taxonomy). This problem arises because datasets for the EE task require specific details about arguments for each new event type added to the taxonomy. However, for the ED task in general, and the OED task in particular, there is no need to consider arguments since the goal is to detect event triggers only. 
Second, when using an existing dataset developed for a particular new task or domain, the event taxonomy and the source of the news articles usually do not suit precisely the needs for the new task or domain. Third, existing datasets do not distinguish ongoing events (which are the focus of our work) from historical, future, hypothetical, or other forms of events. Because of these three disadvantages, we rely on creating our own dataset for our specific task.

The state-of-the-art approach for ED is to generate a neural-based classifier with a class for each possible event type, and an extra class for non-events. Traditionally, these classifiers use a wide variety of features representing lexical, syntactic or entity information, which are typically the result of applying  NLP tools (e.g.~\citep{li2013joint}). With the advent of neural network language models and the availability of distributed continuous representations for words and sentences learned in an unsupervised fashion from large corpora (such as word and sentence embeddings), the features used in the models changed radically. A lot of the input features used in state-of-the-art approaches are unsupervised representations automatically learned from big corpora (e.g., Word2vec~\citep{mikolov2013distributed} and Fasttext~\citep{bojanowski2017enriching}), and many of these representations can keep improving during training through gradient descent. 

According to~\citep{boros2018neural}, ED approaches fall into one of three categories: pattern-based~\citep{krupka1991ge,hobbs1992sri,riloff1996automatically,riloff1996empirical,yangarber2000automatic}, feature-based~\citep{freitag1998information, chieu2003closing, surdeanu2006hybrid, ji2008refining,patwardhan2009unified,liao2010using,huang2011peeling,hong2011using,li2013joint,bronstein2015seed}, and neural-based~\citep{chen2015event,nguyen2015event,nguyen2016joint,nguyen2016two,feng2018language}. The first two rely on sophisticated handcrafted rules, patterns, and features, as well as in NLP tools. The third category relies on neural networks for both feature representation and token prediction. In this paper, we focus on the last category, which solves a lot of the problems with the first two categories mentioned above, while achieving state-of-the-art results for both ED and EE.

 EE systems typically follow one of two possible architectures: pipelined \citep{chen2015event,ji2008refining,liao2010using,hong2011using} or joint \citep{zhang2019joint,liu2018jointly,sha2018jointly}. In the pipelined architecture, the ED task is the first step, in which the trigger word is detected and classified. Afterward, the system performs the rest of the EE task by extracting the arguments for those triggers. In the joint architecture, several proposals employ the joint architecture in which argument extraction is part of the trigger extraction phase and vice versa. 

The notion of context information is not new for the ED task. In \citep{feng2018language}, the authors use an RNN model to capture a simple notion of context for each word to improve the performance on the task while using classical embeddings. In \citep{fan2020adverse}, the authors show how BERT embeddings improve the performance on the adverse drug event (ADE) detection task in the medical domain.  In~\citep{ tong2020image} BERT-based text features are integrated with image features to address the ED task in news. Even though these approaches are either applied to domains different from news or incorporate non-textual information, they provide evidence supporting the usefulness of exploiting contextual word representations for ED.

On the other hand, the notion of ongoing events as defined in this paper ---to the best of our knowledge--- is introduced here for the first time. In \citep{huang2016distinguishing}, the authors use document classification to categorize news articles containing events as past, ongoing, future planned, future alert, and future possible. Although they point to the importance of distinguishing ongoing from past and future events, their approach is different from ours. They work at the document level, classifying full news articles, while we work at the level of event triggers. Furthermore, they classify events into different categories (past, ongoing, future), while we only determine if the event is ongoing or not. Additionally, the dataset they introduce, EventStatus corpus, is defined for a different context than ours. Their dataset consists of civil unrest events, such as protests, demonstrations and strikes.

\section{The Ongoing Event Detection Task}
\label{sec:OEDTask}

The OED task is a specific ED task whose goal is to detect ongoing event mentions only, as opposed to historical, future, hypothetical, or other forms of events that are neither fresh nor current. Current states of affairs reported in the news are also considered ongoing events.

In this section, we present all the definitions required for the OED task, and we present several key examples to facilitate the understanding of the task. All these definitions, the examples for the task, and the annotators' guidelines are available on the dataset website.\footnote{\url{https://cs.uns.edu.ar/~mmaisonnave/resources/ED\_data/}.}

%% DEFINITIONS
\theoremstyle{definition}
\begin{definition}[Ongoing Event]
An ongoing event is any text fragment in a news article reporting a real-world event that meets any of the following conditions: (i) it is a fresh event, (ii) it happened a time ago and is still ongoing, or (iii) it is the current state of affairs for a given entity.
\end{definition}
An example of (i) is when an earthquake just took place, and a news article fragment covers that event. An example of (ii) is when a riot started in a city some days ago and is again reported in the news while it is still happening (because there is some new information or is a recapitulation of what happened). Lastly, an example of (iii) is when a news article reports a crisis or a recession that is taking place in some country or region. It is important to notice that the same fragment could contain more than one ongoing event. For example, some news article fragments could be reporting an event caused by another. For instance, an ongoing crisis (an example of (iii)) could cause a riot (an example of (ii)).

Ongoing Event Trigger and Ongoing Event Detection Task are defined analogously to Event Trigger and Event Detection Task~\citep{walker2006ace}, respectively, as follows:

\begin{definition}[Ongoing Event Trigger]
An Ongoing Event Trigger is the word that most clearly states the occurrence of an ongoing event (Definition 1).
\end{definition}
\begin{definition}[Ongoing Event Detection Task]
The Ongoing Event Detection (OED) Task is the task of detecting the ongoing event trigger (Definition 2).
\end{definition}

In the OED task, the context of a word is crucial to determine if it refers to an ongoing event or not. For example, take the word ``crisis'' in the following sentences:
\begin{enumerate}
\item The current {\em crisis} will accelerate digital technology.
\item There will not be a {\em crisis} in the foreseeable future.
\item The same trend could be observed during the Global Financial {\em Crisis} more than a decade ago.
\item Any financial {\em crisis} is catastrophic, and we must mitigate the risks of a future {\em crisis}.
\end{enumerate}
Only the reference to a ``crisis'' in sentence 1 is considered an event trigger, while the other mentions of the same word  are not. This is guided by the need to distinguish ongoing events from those that are not, which is a requirement of the ultimate goal of our broader project of detecting causal relations between events reported in the news.  Most existing approaches do not adopt this context-sensitive definition of event. Furthermore, our definition of ongoing event also accounts for current states of affairs, which are not taken into consideration by existing proposals.

As another example, consider the following news extract: ``devaluation is not a realistic option to the current account deficit since it would only contribute to weakening the credibility of economic policies as it did during the last crisis.” The only word that is labeled as ongoing event trigger in this example is “deficit” because it is the only ongoing event referred to  in the news. The word “devaluation” is not an ongoing event trigger as a devaluation may not take place. Similarly, the word “weakening” is not an ongoing event trigger as it is a hypothetical event. Finally, the word “crisis’’ is not considered an ongoing event trigger as the news refers to a crisis from the past. Note that the words “devaluation”, “weakening” and “crisis’ could be labeled as ongoing event triggers in other news extracts, where the context of use of these words is different, but not in the given example. 

Additional details on the OED task, including a description of the system used to assist the labeling process can be found on the ``Ongoing Event Detection Task's Annotation Guidelines'' available on the dataset website (\href{https://cs.uns.edu.ar/~mmaisonnave/resources/ED\_data/}{https://cs.uns.edu.ar/$\sim$mmaisonnave/resources/ED\_data/}).

\section{Experimental Setup} 
    \subsection{Dataset} \label{sec:dataset}
To build our dataset we tokenized the full New York Times (NYT) archive (1987-2007) \citep{sandhaus2008new} using the Spacy NLP library and divided the news into sentences. 
From the full set of sentences extracted from the corpus ($\sim$64 million), we selected a subset for labeling. We chose three episodes of real-world crises: the Mexican peso crisis of 1994, the Russian financial crisis of 1998, and the Asian financial crisis of 1997. We set up the search engine Lucene\footnote{\href{https://lucene.apache.org/}{https://lucene.apache.org/}} (with the default configuration) to search sentences related to these three episodes. We performed a search using keywords manually selected by experts. Examples of these keywords include ``Mexico'', ``crisis'', ``debt'', ``capital flight'', and ``devaluation''. From the obtained results, we randomly selected 2,000 sentences. Also, we randomly selected from these results a separate set of 200 sentences for testing purposes.

We chose to use sentences instead of full texts to favor diversity in our dataset. Because the annotation task is labor-intensive, by labeling sentences instead of full texts we covered different events from different news articles rather than repeated mentions of the same event discussed in various parts of the same article. We believe the decision of using sentences instead of full texts does not simplify the task. Instead, it changed the task to a different one, with different advantages and disadvantages. In this context, we have shorter texts, which means we could use simpler models. However, we have the problem that context was missing for some instances (because important information was not part of the sentence being labeled). In that sense, this task is similar to short-text classification.

Since in the proposed OED task we are not limited to a fixed set of events, any ongoing real-world event or situation reported in the news articles was considered an event. Consequently, each word in our dataset was labeled as {\em event trigger} or {\em non-event trigger}, transforming the OED task into a binary classification task. As mentioned earlier, it is important to distinguish those events and situations that are in progress (or are reported as fresh events) at the moment the news is delivered from past events that are simply brought back, future events, hypothetical events, or events that will not take place. In our dataset we only labeled as {\em event trigger} the first type of event. Based on this criterion, some words that are typically considered as events are labeled as {\em non-event triggers} if they do not refer to ongoing events at the time the analyzed news is released. This is illustrated by the example presented in Section~\ref{sec:introduction} with different mentions of the word ``crisis''.

We developed a simple active learning tool to assist the labeling process of the training and validation data (but not the test data). This tool used an early prototype of the RNN model used for event prediction (described in Section \ref{sec:RNNmodel}) to suggest labels. Each sentence was presented to four users for labeling, along with the corresponding suggestions generated by the model. Then the four users had to agree on keeping the suggested labels, removing some of them, or adding new ones. The whole process took a total of fifteen sessions of approximately two hours each. 
Since the labeling process was carried out following a consensus-based approach, no inner-agreement could be calculated. Therefore, such a metric is not reported.
Because we initially do not have any labeled instances for training, we had a {\em cold start}.  As new instances were labeled, the model was re-trained. The re-training process took place each time 50 news instanced were labeled, and each re-training process was performed using the whole set of instances labeled at that moment (including the newly labeled 50 instances).
It is important to notice that to avoid biasing the users' decisions when tagging events in the test set, the four users were presented with sentences with no suggestions provided by the tool and had to reach a consensus on which words had to be marked as event triggers. 
Because the annotation process is so labor-intensive, the size of the test set is relatively small. However, for the results to be reliable, statistical tests were used on the results reported in section \ref{sec:results} to guarantee that the hypotheses tested in this work are valid.

The statistics of the dataset are presented in Tables~\ref{table:dataset_vocab} and~\ref{table:dataset_stats}. Additional details on how the dataset was built can be found on the dataset website (\url{https://cs.uns.edu.ar/~mmaisonnave/resources/ED\_data/}). 

\begin{table}[t]
\begin{center}
\resizebox{0.8\columnwidth}{!} { 
\begin{tabular}{lr}
\hline
\multicolumn{2}{c}{Full Dataset (training/validation + test)} \\ \hline
Sentence Count        & 2,200    \\
Word Vocab. Size       & 8,647    \\
Entity (E) Vocab. Size     & 34       \\
Part-Of-Speech Simplified (P) Tag Vocab. Size     & 16       \\
Dependency Parser (D) Tag Vocab. Size    & 47       \\
Part-Of-Speech Detailed (T) Tag Vocab. Size    & 47       \\ \hline
\end{tabular}
}
\caption{Statistics about the OED dataset vocabulary.}  \label{table:dataset_vocab}

\end{center}
\end{table}

\begin{table}[b]
\resizebox{\columnwidth}{!}{%
%@{\extracolsep{4pt}}lcccccc@{}
\begin{tabular}{@{\extracolsep{4pt}}lcccc@{}}%{lcccccc}
\hline
\multirow{2}{*}{Metric} & \multicolumn{2}{c}{Training/Validation} & \multicolumn{2}{c}{Test}  \\ \cline{2-3} \cline{4-5} 
                      & Total    & Avg. per Sent   & Total & Avg. per Sent      \\ \hline

Token Count  & 76,629 & 38.31 & 7,382 & 36.91  \\ 
Word Count   & 67,032 & 33.52 & 6,442 & 32.21  \\
Entity Count & 11,502 & 5.75  & 950   & 4.75   \\
Event Count  & 5,119  & 2.56  & 416   & 2.08    \\
\hline
\end{tabular}
}
\caption{Total number of tokens, words, entities, and events found in the  dataset.}  \label{table:dataset_stats}

\end{table}

\subsection{RNN Model} \label{sec:RNNmodel}
\label{sec:rnn}
We designed and trained a Recurrent Neural Network (RNN) for the OED task. We conducted several experiments combining different hyperparameters, features, and architectures. In this section, we will review the different configurations tested and the intuition behind each selection.

\smallskip
\textbf{Features.}
To be able to detect ongoing events in natural language text, we hypothesize that syntactic, semantic, and grammatical information is needed. To represent the semantics of each token we use  Word2vec \citep{mikolov2013efficient} ($W$). Note that we use Word2vec instead of other word embeddings (e.g., FastText \citep{bojanowski2017enriching}) for the sake of comparison, as Word2vec is used in our baseline \citep{nguyen2015event}.
 Word2vec embeddings represent each word by means of a 300-dimension vector that is fine-tuned as the model is trained for OED. We also tested context-sensitive tensors provided by the Spacy Library ($Sp$). A Spacy context-sensitive tensor is a 96-dimension vector, which is the internal state of the neural model used for NLP by the Spacy library\footnote{\href{https://spacy.io/}{https://spacy.io/}}. This feature was not fine-tuned during training.

To encode the syntactic information, we first used the Spacy library to identify the dependency tree ($D$). For each token, we extracted the dependency relation with the head token in the dependency tree. Using that information, we trained a 10-dimension supervised Keras embedding layer. This layer was initialized with random weights and the representation was incrementally fine-tuned along with the training of the rest of the network. The grammatical information was encoded using two different embeddings representing information retrieved through Spacy's Part-Of-Speech (POS) tagger. We used both the simplified POS tag version ($P$) and the detailed one ($T$) of the Spacy NLP library. As we did in the case of the syntactic information, we used a first embedding layer to transform these two categorical variables into two vectors of 10 dimensions each and fine-tuned the representation during training.

We used the Spacy Named Entity Recognizer tagger to include entity information for the OED task in the form of a separate feature ($E$). We represented each word as a two-part tag. The first part is the IOB notation and the second is the type of the entity. As we did for other features ($T$, $P$, $D$), we used an embedding layer to transform a one-hot encoding of this categorical input into a 10-dimension vector. This layer is initialized with random weights and is fine-tuned during training.

Contextual embeddings, such as ELMo \citep{peters2018deep} and BERT \citep{devlin2018bert}, were proposed to solve the problem of mixed semantics when the same word has a context-sensitive meaning. These embeddings achieve ground-breaking performance on a wide range of NLP tasks. We hypothesize that having vector sensitivity to the context is crucial for the OED task, and it will bring a boost in performance. To illustrate this hypothesis, consider the following two sentences: ``The firm had to fire employees'' and ``Fire burns a home near Brainerd airport.'' They both use the word ``fire'', written in the same way, but with totally different meanings. Word2vec will only have one vector for this word, which can limit the performance of an ED model for identifying these cases.  Motivated by this intuition, we incorporated pre-trained contextual word embeddings, in particular BERT embeddings ($B$), as an input to the model.  

Another problematic situation occurs with a word that preserves its semantics across different sentences but that can refer to an ongoing event trigger or not, depending on its context of use. This was illustrated in Section~\ref{sec:introduction} with the word ``crisis’’, which can be used to refer to a recent or ongoing event (event trigger) or to a historical, future, hypothetical or other forms of events that are neither fresh nor current (non-event triggers). Following this intuition, we incorporate for each token a 768-dimensional vector feature representing a contextual embedding for the whole sentence. 
Note that each token in the same sentence will have the same contextual sentence embedding as input. The intuition behind using a common contextual sentence embedding for each token in a sentence is that sentences usually give information either about past, future and hypothetical situations (which were irrelevant to us), or about current events (which are the ones that we want to detect). It was  uncommon for a sentence to mix these two types of information.  Guided by this intuition, we incorporate  a contextual sentence embedding ($S$) built by adding up the BERT embeddings for each token in the sentence.
The $B$ and $S$ inputs are not modeled with a Keras layer and therefore they are fine-tuned during training for the OED task.
A summary of the eight above-described features used for the RNN models is presented in Table~\ref{table:RNN_features}.

\begin{table}[t]
\resizebox{\columnwidth}{!}{%

\begin{tabular}{ccl}
\hline
%\multicolumn{3}{c}{Features}                                                       \\ \hline
Abbr. & Size & Description                                                                \\ \hline
$W$     & 300 & Pre-trained word embeddings. \\% Word2vec.                                      \\
$P$     & 10 & Part-Of-Speech tag embeddings, simplified version.                          \\
$T$     & 10 & Part-Of-Speech tag embeddings, detailed version.                                \\
$D$     & 10 & Dependency parser tag embeddings.                                           \\ 

$E$     & 10 & Entity tag embeddings. \\ %Computed using the Spacy NER tagger. \\

$Sp$    & 96 & Spacy contextual word embeddings.                             \\
$B$     & 768 & Pre-trained contextual word embeddings.\\ %BERT embeddings                         \\
$S$     & 768 & Contextual sentence embeddings.\\% Computed adding the B embeddings. \\

\hline
\end{tabular}
}
\caption{{\bf Features} used by the RNN models.}  \label{table:RNN_features}

\end{table}

We implicitly use transfer learning in the various inputs of the model. Firstly, by using the pre-trained Word2vec, Spacy and BERT embeddings, we are incorporating semantic information extracted from large corpora of unlabeled text. Secondly, by using the POS tagger and the Dependency tagger from the Spacy library, we are including grammar and syntactic information coming from other IE fields, where there are large labeled corpora.

\begin{figure}[!ht]
    \centering
    \includegraphics[width=\textwidth]{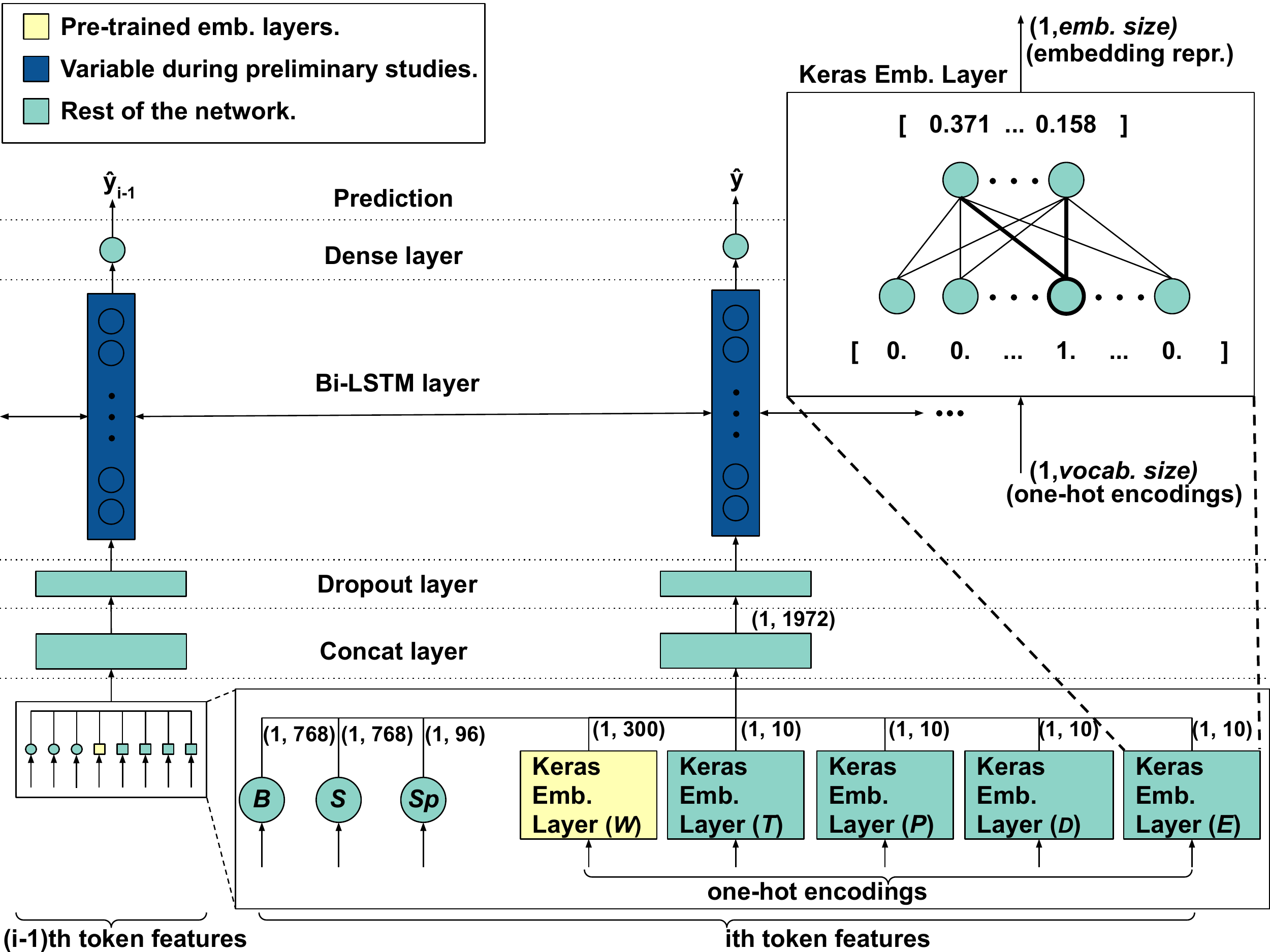}
    \caption{\textbf{Outline of the proposed RNN architecture}. From bottom to top. A representation of the eight inputs, for the $i$th and the $(i-1)$th token. For each token, three of the inputs are vector representations ($B$, $S$, $Sp$); the other five are one-hot encodings that enter five Keras embedding layers. Each of thee dense layers is depicted in the top-right corner diagram. One of these embedding layers start with pre-trained weights ($W$), the others with random weights. After the embedding layer, the eight vectors are concatenated together in a single 1972-dimensional vector. This vector enters a dropout layer. The output of the dropout layer enters a  Bi-LSTM layer with fifteen hidden units. During the preliminary studies, we tested different numbers of Bi-LSTM layers and hidden units. For the held-out data, we use the configuration depicted in this Figure. The output of the Bi-LSTM layer is the input to a last dense layer, which makes the prediction.}
    \label{fig:RNN_diagram}
\end{figure}

\smallskip
\textbf{Architecture.}
We chose an RNN architecture based on Long-Short Term Memory (LSTM) cells to exploit the dependencies between previous and subsequent tokens for the classification of the current word. The inputs for each word, which are eight embeddings, are all concatenated to form a vector of 1962 dimensions.  We added a Dropout layer after the concatenated embeddings.
Following the dropout layer, we added a Bidirectional LSTM (Bi-LSTM) layer with 15 hidden units. A final dense layer with only one hidden unit was added to the Bi-LSTM Layer. The output of this final layer is the prediction. The RNN architecture is outlined in Figure \ref{fig:RNN_diagram}. Although this architecture is the one used during the experiments on the data held-out for testing, we tested other architectures (different numbers of Bi-LSTM layers and hidden units). The results of these additional experiments with various architectures can be found in the annexes.

\smallskip
\textbf{Hyperparameters.}
For the embedding layers we use the default configuration, only changing the random initialization for the Word2vec embeddings. 
We use $p=0.1$ for the Dropout layer and we set up every Bi-LSTM using the default configuration, adding only L1L2 regularization with values of 0.001 for both L1 and L2. For the final experiments on the data held-out for testing, we used a single Bi-LSTM layer architecture with 15 hidden units, which was the architecture with the best performance during the preliminary studies reported in the annexes.
Finally, we set up the dense layer using the sigmoid activation function.

\subsection{Baseline Models} \label{sec:baseline}
 Since no previous model was validated using our dataset, existing metrics and results from the ED field are not directly comparable to ours. For this reason, we replicated an ED model from the state-of-the-art to use as a baseline, as well as a classical SVM model. Because the focus of our work is ED (and more specifically OED), we evaluated our model against state-of-the-art in ED, instead of considering approaches that perform ED+EE. Based on these considerations and because of its simplicity and comparable performance with the state-of-the-art in ED, we selected and replicated the Convolutional Neural Network (CNN) model proposed in \citep{nguyen2015event} as a baseline for comparison (in addition to the SVM baseline model).
Despite being proposed several years ago, this model is still competitive with the available state-of-the-art methods. For example, it presents an F1-score of only 4.1 percentage points lower than more recent models,  such as \citep{nguyen2018graph}. The selected baseline reports an F1-score of 69\% for the ED task using gold-standard entity annotations, while the model from \citep{nguyen2018graph} reports an F1-score of 73.1\% on the same dataset (ACE 2005 Corpus) and using the same gold-standard entity annotations. The authors do not report in~\citep{nguyen2018graph} the result of the more recent neural model for predicted entities. However, they do report in~\citep{nguyen2015event} the F1-score for predicted entities for the model we use as our baseline. For this setting, the model achieves an F1-score of 67.6\%. 

As mentioned earlier, in order to compare our work with a classical approach, we also implemented a SVM-based ED model using the Scikit-learn library. The SVM approach used Word2vec embeddings as features ($W$) and it was tested with four different kernels and default parameters.

The  CNN model from \citep{nguyen2015event} was replicated as faithfully as possible. However, although we tried to replicate the model exactly as it is in the original paper, to adapt it to our dataset, we had to make some minor changes. Furthermore, since the code was not available, some minor implementation details may not be the same as in the original model. We had to make some decisions about some aspects and configuration of the models which were not explicit in the original paper. For example, we had to decide which NLP tools to use for entity extraction. In our work, we use the Spacy library for all the NLP related tasks.
In the remainder of this section, we describe in detail the CNN model used. We also describe some minor changes and decisions that we had to make, explaining the rationale behind them.

\textbf{CNN features.}
As in the original paper, we use three input features for the baseline model. First, we use the Word2vec embeddings as in our RNN model ($W$). Second, we use entities embeddings ($E$). For building these features, we use the Spacy Named Entity Recognition (NER) tagger and a Keras embedding layer for building the embeddings, using the former to get categorical variables (tags) and the latter for transforming these variables into vectors. We chose Spacy as our NER tagger, as in the original paper there is no mention of a specific tool, and because of the state-of-the-art performance of the Spacy library in several NLP tools. The last feature employed for the baseline model is a position embedding ($Po$), which represents the relative position of each word with respect to the current token under classification. The word embedding layer starts with a representation learned from the pre-trained Word2vec vectors, while the other two start with random weights, as in \citep{nguyen2015event}. The three layers are updated while training for the ED task. 

\begin{figure}
    \centering
    \includegraphics[width=\textwidth]{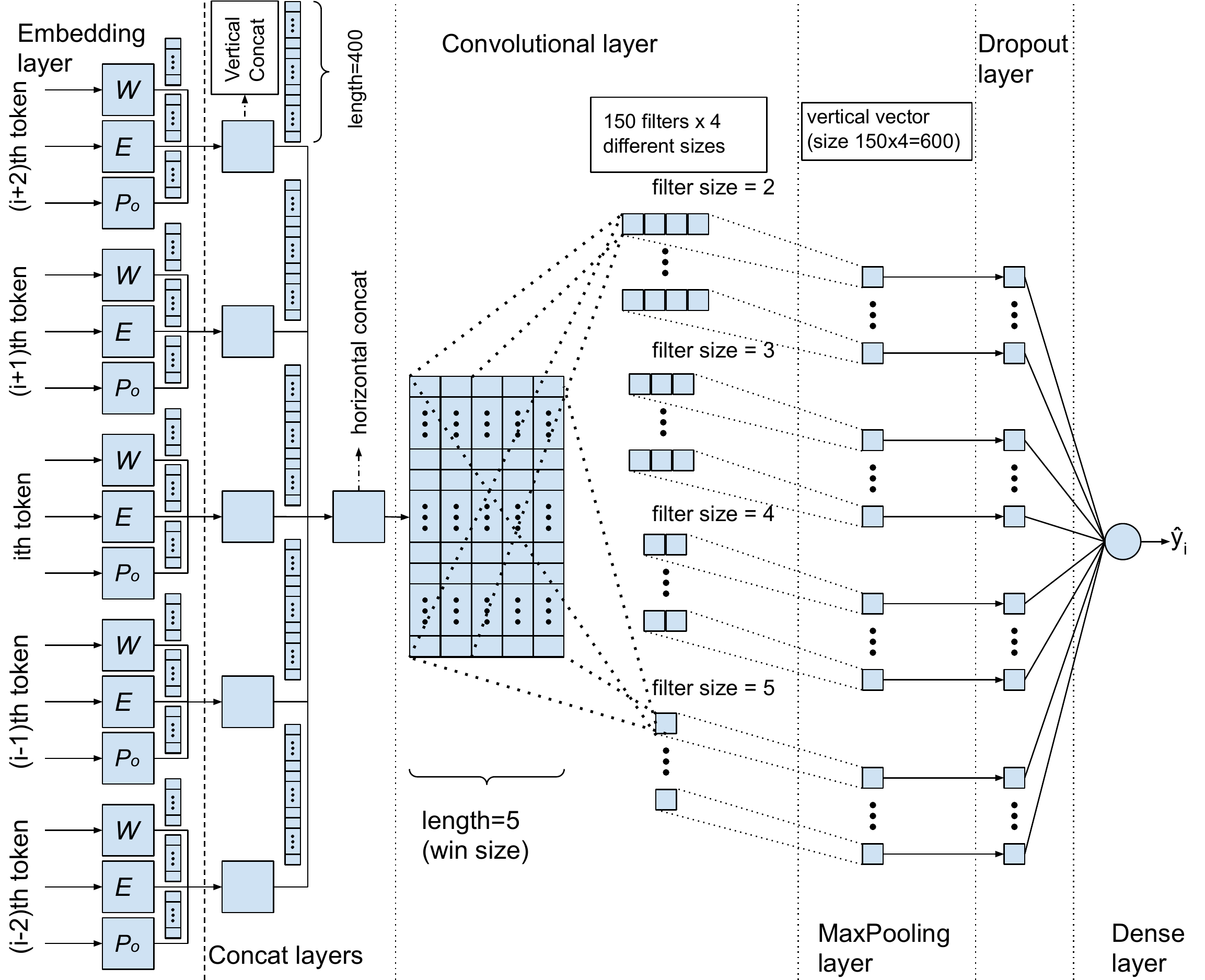}
    \caption{\textbf{Outline of the baseline model (CNN)} with window size 5. Descriptions are given from left to right. (a) Inputs. A representation of the three different inputs (Word embedding $W$, Entity embedding $E$ and Position Embedding $Po$) for the five tokens in the window. Each input is a one-hot encoding. (b) Embedding layer. Input enters a Keras embedding layer. (c) Concatenation layers. We concatenated the resulting embeddings in a single 400-dimension vector that represents each token. Another concatenation layer stacks these five vectors into one matrix. (d) A convolutional layer applies 600 filters of four different sizes (2, 3, 4, and 5) with 150 filters for each size. (e) MaxPooling layer. A max-pooling layer is applied afterward. (f) Dropout and Dense layers. Finally, the last two layers are a dropout and a dense layer. The dense layer is the last one of the network and is the one that makes the prediction.}
    \label{fig:CNN_diagram}
\end{figure}

\textbf{CNN architecture.}
The architecture used in \citep{nguyen2015event} is a one layer CNN, followed by a max-pooling layer and lastly a dropout layer followed by a dense layer for the prediction. The inputs of this network are three lookup tables, with the three types of embeddings. The representation in theses lookup tables improves along with the training for the ED task. We replicated the same architecture and behavior, by replacing the lookup tables with embedding layers that meet the same purpose: storing and providing a vectorized representation, and improving the representation while training for the ED task. The remaining of the network follows the same architecture as in \citep{nguyen2015event}. An outline of the architecture of the CNN model is presented in Figure \ref{fig:CNN_diagram}.

\textbf{CNN hyperparameters.}
While many of the hyperparameters remain the same, since our data is different from the one used in \citep{nguyen2015event}, the window size used had to be changed to better suit the data. Since in our dataset, each data item is a sentence and not a whole document like in ACE 2005 (the dataset used in \citep{nguyen2015event}),  the window sizes had to be adjusted. We tried with smaller window sizes, 1, 3, 5, 11, 21, and also for comparison sake we tested the model with window size 31. We use the same number of filters (150), and the same size for the filters (2, 3, 4, 5), as in the original paper.  We used the sigmoid activation function for the final dense layer to use the same performance metrics as in our RNN model. We use, as in the original paper, a batch size of 50, a probability for the dropout layer of 0.5, and we set the hyperparameter for the l2 norms  to 3. We used the binary Cross-Entropy loss function and Adam's optimizer.

\section{Results and Discussion} \label{sec:results}
 In this section, we present the results and discussions for  different variations of the proposed (RNNs) and baseline models (CNNs and SVM). For each variation and each model (excluding the SVM models), we randomly split the training/validation part of the dataset into different training and validation subsets. For each model variation, we trained the model until it did not improve the F1-score on the validation set for 400 epochs (early stopping with patience 400). We used consecutive seeds from 1 to 10 to guarantee replicability. The model with the best performance on the validation set was selected and used to make the predictions on the test set. In the case of the SVM models, since early stopping was not necessary, we use the training and validation sets together (training+validation) for training. Because the SVM model is not stochastic, only one run on the training+validation and test data was executed for each of the four kernels. Therefore, the performance values reported for the SVM models are the results from a single run rather than average values.
 
 It is important to notice that we used the same held-out data to test all the models and this data was never used during the training stage. In this section, we present and discuss the performance of each model on the validation splits and test data. We report results for a total of fifteen different model variations. For selecting these model variations, we run several preliminary studies. The preliminary studies and their discussion are presented in the annexes. For each variation of the neural-network models, we report the average metrics for the ten trials, which include the average sensitivity, specificity, and harmonic mean between these two metrics, namely F1-score. We do not report the average accuracy for the model, because, in the presence of highly unbalanced data (93.41\% are non-events), the accuracy is a misleading metric.  To thoroughly analyze the F1-score metric on the test set we also report the confidence intervals (CI) at 95\% level of confidence and the p-value for a t-test between the model considered and the best model of each table. For example, in the first row of Table~\ref{table:RNN_final}, we report the p-value of a single-tail t-test between the F1-score values achieved by the first model (model 1) and the best model of the table (model 7). We used the F1-score metric on the test data for the t-test.
Since the SVM model is not stochastic, only one run on the training+validation and test data was executed for each of the four kernels. Consequently the results reported are from a single run, and therefore no t-test could be performed.

\subsection{Experiments on the Test Set}
\label{sec:experiments}
\textbf{For the proposed RNN model}, we run ten trials for seven different sets of features (models 1 to 7) as depicted in Table~\ref{table:RNN_final}.  Although we tried different numbers of Bi-LSTM layers and hidden units, we only evaluated on the test set the architecture that achieved the best performance during the preliminary studies, i.e., the architecture with a single Bi-LSTM layer with 15 hidden units. The  results are reported in Table~\ref{table:RNN_final}.

We conducted these experiments for two main reasons. First, to test the hypothesis that the contextual word embeddings ($B$) are having a significant positive impact on the overall performance. We tested this hypothesis during the preliminary studies, and we found further evidence during the experiments on the test dataset. The significant drop in performance (of 11.7 percentage points) from including or excluding the contextual word embeddings (model 1 vs. model 2) is indicating the importance of these features for the OED task. 
Second, we carried out the rest of the experiments (models 3 to 7) as an ablation study to assess the impact of the proposed features to the overall performance. We measure the contribution of each feature during the preliminary studies. During the experiments on the test dataset, we found further evidence that several inputs are not contributing to the performance in the presence of the other features. These results allowed us to simplify the model and obtain even better results. 
 Model 7, which excludes all the attributes except the contextual word and sentence embeddings, is the model with the best performance. This result shows that the other features are not relevant in the presence of these two other features, and they were adding noise and complexity to the model. The statistical analysis between model 7 and the others shows that all the differences are significant at the level of 90\% (p-values are smaller than 0.1).

\setlength{\tabcolsep}{3pt} % Default value: 6pt

\begin{table}[t]
\resizebox{\columnwidth}{!}{%
\begin{tabular}{ccccccccccl}
\hline
\multirow{2}{*}{Model}  & \multirow{2}{*}{Features} & \multicolumn{3}{c}{validation} & \multicolumn{5}{c}{test} \\ \cmidrule(r){3-5} \cmidrule(l){6-10}
    &            & $\overline{sens}$ & $\overline{spec}$ & $\overline{F1}$     & $\overline{sens}$ & $\overline{spec}$ & $\overline{F1}$ & F1 CI  & p-value          \\ \hline
1   &   all          &  	    0.696   & 	0.945       & 	0.712               & 	0.667       & 	0.931   & 	0.676           & 	    $\pm$ 0.026     & 0.024 \\ \relax
2   &   all-\{$B$\}                    & 	    0.545   & 	0.957       & 	0.591               & 	0.522       & 	0.948   & 	0.559           & 	 	$\pm$ 0.046     & 0.000 \\ \relax
3   &   all-\{$T$,$P$,$D$\}            &	    0.677   &   0.949       & 	0.697               & 	0.654       & 	0.935   & 	0.666           &       $\pm$ 0.031     & 0.012 \\ \relax
4   &   all-\{$T$,$P$,$D$,$W$\}          &	    0.703   & 	0.948       & 	0.718               & 	0.686       & 	0.935   & 	0.690           & 	 	$\pm$ 0.018     & 0.076 \\ \relax 
5   &   all-\{$T$,$P$,$D$,$Sp$\}         &	    0.689   &   0.945       &   0.706               &   0.672       &   0.931   &   0.683           &       $\pm$ 0.013     & 0.007 \\ \relax
6   &   all-\{$T$,$P$,$D$,$E$\}            &       0.686   &   0.949       &   0.706               &   0.662       &   0.936   &   0.675           &       $\pm$ 0.023     & 0.012 \\ \relax
7   &   all-\{$T$,$P$,$D$,$Sp$,$W$,$E$\}     &	0.726   &   0.943       &     \textbf{0.734}    &   0.706       &   0.928   &   \textbf{0.704}  &    	$\pm$ 0.012     & ---\\ \hline

\end{tabular}
}
\caption{Average performance of the {\bf RNN}  model in the validation and test set for seven different sets of features using an architecture with a single  Bi-LSTM layer and 15 hidden units. We describe the features used by each model by indicating which features are removed from the full set of features (all), where the full set of features is \{$T$,$P$,$D$,$Sp$,$W$,$E$,$B$,$S$\}. We adopted this notation to simplify our experiments' interpretation as an ablation study (comparing the model with all the features against the models with some features removed). The training set is used for training and the validation set for early stopping, while the test set is held out for testing. The resulting average performance during training is omitted from the reported analysis.   } \label{table:RNN_final}

\end{table}

\textbf{The baseline CNN model} was run for ten trials with  different window sizes. Here, we analyze windows of size 1 and 11 (other window sizes are examined in the annexes). For both window sizes, we run experiments with and without the entity embedding, but always including the word embeddings. For window size 1, we excluded the position embeddings. The results of these experiments are presented in Table~\ref{table:CNN_final}.
 The best CNN baseline (model 8) shows an F1-score of 0.575 on the test data, which is significantly lower than the performance of all the proposed RNN models, except for the one that excludes the contextual word embeddings (model 2).

\begin{table}[b]
\resizebox{\columnwidth}{!}{%
\begin{tabular}{ccccccccccl}
\hline
\multirow{2}{*}{Model} & \multirow{2}{*}{Win Size} & \multirow{2}{*}{Features} & \multicolumn{3}{c}{validation} & \multicolumn{5}{c}{test} \\ \cmidrule(r){4-6} \cmidrule(l){7-11}
     &&          & $\overline{sens}$ & $\overline{spec}$ & $\overline{F1}$     & $\overline{sens}$ & $\overline{spec}$ & $\overline{F1}$ & F1 CI  & p-value          \\ \hline
    
8 & 1           & \{$W$,$E$\}        & 0.477 & 	0.971 & 	0.570 & 	0.499 & 	0.968 & 	\textbf{0.575}           &  	$\pm$ 0.031 & 	---         \\
9 & 1           & \{$W$\}          & 0.472 & 	0.972 & 	0.565 & 	0.493 & 	0.969 & 	0.569  & 	$\pm$ 0.031 & 	0.384	        \\

10 & 11	        & \{$W$,$E$,$Po$\}	     & 0.507 & 	0.963 & 	0.596 & 	0.326 & 	0.960 & 	0.394           & 	$\pm$ 0.028 & 	0.000	    \\
11 &	11	        & \{$W$,$Po$\}	     & 0.499 & 	0.965 & 	0.589 & 	0.345 & 	0.942 & 	0.406           & 	$\pm$ 0.035 & 	0.000 \\

\hline
\end{tabular}
}
\caption{ Average performance of variations of the {\bf CNN} model on the validation and test sets. The training set is used for training and the validation for early stopping, while the test set is held out for testing. The average performance during training is omitted from the reported analysis.   }  \label{table:CNN_final}
\end{table}

To compare the RNN models and the best CNN baselines, we evaluated the p-values of single-tail t-tests between the average F1-score for the worst and the best RNN models (models 2 and 7, respectively)  against the two best CNN models (models 8 and 9).
 Model 7 is statistically significantly better than all the others having a p-value of 0.0 for all the t-tests.  
The t-test between models 8 and 9 has a p-value of 0.384.  This high p-value indicates that the difference in performance of these two models is not statistically significant, which points to the fact the entity embeddings ($E$) have a negligent impact on performance. The difference in performance of the worst RNN (model 2) with respect to the best CNNs (models 8 and 9) is not statistically significant either (p-value of 0.264 and 0.344, respectively).

\textbf{The baseline SVM model} was run on the training and test data with linear, polynomial, RBF and sigmoid kernels. 
 The results of these experiments are presented in~Table \ref{table:SVM_final}.
 The best SVM baseline  (model 15) shows an F1-score of 0.564 on the test data, which is significantly lower than the performance of all the RNN models except for the one that excludes the contextual word embeddings (model 2). On the other hand, its performance is only slightly inferior to that achieved by the two best CNN baselines  (models 8 and 9).

\begin{table}[t]
\centering
\resizebox{0.8\columnwidth}{!}{
\begin{tabular}{cccccccl}
\hline
\multirow{2}{*}{Model} & \multirow{2}{*}{Kernel}  & \multicolumn{3}{c}{training + validation} & \multicolumn{3}{c}{test} \\ \cmidrule(r){3-5} \cmidrule(l){6-8}
     &          & $sens$ & $spec$ & $F1$     & $sens$ & $spec$ & $F1$         \\ \hline

12 & linear       & 0.247         & 0.995         & 0.395         & 0.231         & 0.994         & 0.375  \\
13 & polynomial         & 0.494         & 0.993         & 0.660               & 0.380         & 0.980         & 0.548 \\
14 & RBF          & 0.459         & 0.993         & 0.628                  & 0.346         & 0.984         & 0.512          \\
15 & sigmoid      & 0.372         & 0.957         & 0.536               & 0.401         & 0.949         & \textbf{0.564}       \\  
\hline
\end{tabular}
}
\caption{ Performance of variations of the  {\bf SVM} model on the training+validation and test sets. Because SVM has a deterministic output and there is no need for early stopping, there was no need to use the training and validation tests separately. Therefore, unlike the RNN and CNN models, the SVM models were trained using the training and validation sets together (training+validation set). }  \label{table:SVM_final}
\end{table}

\subsection{Discussion} \label{sec:discussion}
In this section, we review and discuss the behavior of five representative models.
The models selected for this discussion are model 2 (RNN with all the features except for the contextual word embeddings), model 7 (best RNN model), models 8  (best CNN model), and model 15 (best SVM model). 

We hypothesize that the contextual embeddings are a crucial factor for boosting the performance of the proposed models. Therefore, in model 2, we exclude only the contextual word embeddings to assess the impact of them on the performance. Since this model lacks the contextual word embedding, we expected a poor performance in comparison to model 7 (which has the contextual word and sentence embeddings). The poor performance of model 2, which is lower than the best baseline models, provides evidence to support our hypothesis. 
Moreover, the superior performance achieved by model 7 (containing only the contextual embeddings $B$ and $S$ as features) indicates that the inclusion of the other features ($W$, $Sp$, $T$, $P$, $D$, $E$) are not useful for the OED task once the contextual embeddings are  included. It is also interesting to observe that model 15, which has a $sigmoid$ kernel, obtained an F1-score of 0.564 on the test set, outperforming some of the neural-based models (models 2, 10 and 11). 

We draw two major conclusions from the discussed results. First,  in an OED task trained on a small labeled datasets, the use of pre-trained features is crucial. This  follows from the fact that BERT-based contextual embeddings ($B$ and $S$) and  Word2vec embeddings ($W$) had a major impact on the performance of models 7 and 15, respectively. This provides further evidence of the well-known fact that the use of transfer learning in small data settings contributes significantly to improving performance.   Second, in the presence of a small labeled dataset, the choice of features, and in particular the use of transfer learning, is more important than having a large or complex model. This follows from the fact that a classical approach (model 15) that relies on pre-trained features outperformed or had  a performance comparable to that of some of the more complex neural-network models.

\section{Conclusions} \label{sec:conclusions}
The main contribution of this work is the definition of the OED task and an extensive evaluation of an OED framework that combines several features. The best of the proposed models, based on an RNN architecture and BERT-based contextual word and  sentence embeddings, shows an improvement of 13.3\% in F1-score with respect to the best baseline model on the test data. Considering that some more recent approaches for ED are slightly above our baseline (4.1\% for \citep{nguyen2018graph}), we have enough evidence to believe that our model achieves a performance competitive to the state-of-the-art, even outperforming some of the most advanced models. It is also worth mentioning that the code and data for replicating our model are fully available.    

Two main conclusions follow from our extensive analysis. First,  contextual embeddings are better suited for the OED task than the other analyzed embeddings and features. In particular, some features proved to have a negligible impact on performance, in the presence of other features. For example, grammar information (captured using POS and Dependency Parser taggers), entity tags and non-contextual word embeddings showed to have no positive impact on performance as long as contextual embeddings are accounted for in the model.  The absence of a positive impact is not because these features do not carry useful information, but probably because this information is already provided by contextual embeddings. Also, the analysis of a classical baseline model showed that, in this context of small data, it is sometimes more important to have useful features than to have a complex or large model. 

The second conclusion we derive from these results is that although the proposed RNN models (which are best suited for text processing) show more flexibility and effectiveness for the OED task than the analyzed CNN models (which are best suited for computer vision tasks), the major difference in performance between the evaluated models relies on the features used, in particular the BERT-based contextual embeddings. 

Another important contribution is the construction of a public dataset for the OED task. The labeling of the dataset was assisted by an active learning tool. The resulting dataset is particularly useful for the OED task as it focuses on ongoing events only. It also differs from other ED and EE datasets in being independent of a fixed event type taxonomy.

 As part of our future work, we plan to evaluate the impact of contextual embeddings on other ED settings (using other datasets and other models).

\bibliographystyle{apalike}

\begin{thebibliography}{}

\bibitem[Adedoyin-Olowe et~al., 2016]{ADEDOYINOLOWE2016351}
Adedoyin-Olowe, M., Gaber, M.~M., Dancausa, C.~M., Stahl, F., and Gomes, J.~B.
  (2016).
\newblock A rule dynamics approach to event detection in twitter with its
  application to sports and politics.
\newblock {\em Expert Systems with Applications}, 55:351 -- 360.

\bibitem[Ahn, 2006]{ahn2006stages}
Ahn, D. (2006).
\newblock The stages of event extraction.
\newblock In {\em Proceedings of the Workshop on Annotating and Reasoning about
  Time and Events}, pages 1--8.

\bibitem[Bojanowski et~al., 2017]{bojanowski2017enriching}
Bojanowski, P., Grave, E., Joulin, A., and Mikolov, T. (2017).
\newblock Enriching word vectors with subword information.
\newblock {\em Transactions of the Association for Computational Linguistics},
  5:135--146.

\bibitem[Boro\c{s}, 2018]{boros2018neural}
Boro\c{s}, E. (2018).
\newblock {\em Neural Methods for Event Extraction}.
\newblock PhD thesis, Universit\'{e} Paris-Saclay.

\bibitem[Bronstein et~al., 2015]{bronstein2015seed}
Bronstein, O., Dagan, I., Li, Q., Ji, H., and Frank, A. (2015).
\newblock Seed-based event trigger labeling: How far can event descriptions get
  us?
\newblock In {\em Proceedings of the 53rd Annual Meeting of the Association for
  Computational Linguistics and the 7th International Joint Conference on
  Natural Language Processing (Volume 2: Short Papers)}, pages 372--376.

\bibitem[Chen et~al., 2015]{chen2015event}
Chen, Y., Xu, L., Liu, K., Zeng, D., and Zhao, J. (2015).
\newblock Event extraction via dynamic multi-pooling convolutional neural
  networks.
\newblock In {\em Proceedings of the 53rd Annual Meeting of the Association for
  Computational Linguistics and the 7th International Joint Conference on
  Natural Language Processing (Volume 1: Long Papers)}, pages 167--176,
  Beijing, China. Association for Computational Linguistics.

\bibitem[Chieu et~al., 2003]{chieu2003closing}
Chieu, H.~L., Ng, H.~T., and Lee, Y.~K. (2003).
\newblock Closing the gap: Learning-based information extraction rivaling
  knowledge-engineering methods.
\newblock In {\em Proceedings of the 41st Annual Meeting on Association for
  Computational Linguistics-Volume 1}, pages 216--223. Association for
  Computational Linguistics.

\bibitem[Devlin et~al., 2018]{devlin2018bert}
Devlin, J., Chang, M.-W., Lee, K., and Toutanova, K. (2018).
\newblock Bert: Pre-training of deep bidirectional transformers for language
  understanding.
\newblock {\em arXiv preprint arXiv:1810.04805}.

\bibitem[Doddington et~al., 2004]{doddington2004automatic}
Doddington, G.~R., Mitchell, A., Przybocki, M.~A., Ramshaw, L.~A., Strassel,
  S.~M., and Weischedel, R.~M. (2004).
\newblock The automatic content extraction (ace) program-tasks, data, and
  evaluation.
\newblock In {\em Lrec}, volume~2, page~1. Lisbon.

\bibitem[Duan et~al., 2017]{duan2017exploiting}
Duan, S., He, R., and Zhao, W. (2017).
\newblock Exploiting document level information to improve event detection via
  recurrent neural networks.
\newblock In {\em Proceedings of the Eighth International Joint Conference on
  Natural Language Processing (Volume 1: Long Papers)}, pages 352--361.

\bibitem[Fan et~al., 2020]{fan2020adverse}
Fan, B., Fan, W., Smith, C., and Garner, H. (2020).
\newblock Adverse drug event detection and extraction from open data: A deep
  learning approach.
\newblock {\em Information Processing \& Management}, 57(1):102131.

\bibitem[Feng et~al., 2018]{feng2018language}
Feng, X., Qin, B., and Liu, T. (2018).
\newblock A language-independent neural network for event detection.
\newblock {\em Science China Information Sciences}, 61(9):092106.

\bibitem[Freitag, 1998]{freitag1998information}
Freitag, D. (1998).
\newblock Information extraction from html: Application of a general machine
  learning approach.
\newblock In {\em AAAI/IAAI}, pages 517--523.

\bibitem[Granger, 1969]{granger1969investigating}
Granger, C.~W. (1969).
\newblock Investigating causal relations by econometric models and
  cross-spectral methods.
\newblock {\em Econometrica: journal of the Econometric Society}, pages
  424--438.

\bibitem[Hobbs et~al., 1992]{hobbs1992sri}
Hobbs, J.~R., Appelt, D., Tyson, M., Bear, J., and Israel, D. (1992).
\newblock Sri international: Description of the fastus system used for muc-4.
\newblock Technical report, SRI INTERNATIONAL MENLO PARK CA.

\bibitem[Hong et~al., 2011]{hong2011using}
Hong, Y., Zhang, J., Ma, B., Yao, J., Zhou, G., and Zhu, Q. (2011).
\newblock Using cross-entity inference to improve event extraction.
\newblock In {\em Proceedings of the 49th Annual Meeting of the Association for
  Computational Linguistics: Human Language Technologies-Volume 1}, pages
  1127--1136. Association for Computational Linguistics.

\bibitem[Huang et~al., 2017]{huang2017zero}
Huang, L., Ji, H., Cho, K., and Voss, C.~R. (2017).
\newblock Zero-shot transfer learning for event extraction.
\newblock {\em arXiv preprint arXiv:1707.01066}.

\bibitem[Huang et~al., 2016]{huang2016distinguishing}
Huang, R., Cases, I., Jurafsky, D., Condoravdi, C., and Riloff, E. (2016).
\newblock Distinguishing past, on-going, and future events: The eventstatus
  corpus.
\newblock In {\em Proceedings of the 2016 Conference on Empirical Methods in
  Natural Language Processing}, pages 44--54.

\bibitem[Huang and Riloff, 2011]{huang2011peeling}
Huang, R. and Riloff, E. (2011).
\newblock Peeling back the layers: detecting event role fillers in secondary
  contexts.
\newblock In {\em Proceedings of the 49th Annual Meeting of the Association for
  Computational Linguistics: Human Language Technologies-Volume 1}, pages
  1137--1147. Association for Computational Linguistics.

\bibitem[Jacobs et~al., 2018]{jacobs2018economic}
Jacobs, G., Lefever, E., and Hoste, V. (2018).
\newblock Economic event detection in company-specific news text.
\newblock In {\em Proceedings of the First Workshop on Economics and Natural
  Language Processing}, pages 1--10.

\bibitem[Jagannatha and Yu, 2016]{jagannatha2016bidirectional}
Jagannatha, A.~N. and Yu, H. (2016).
\newblock Bidirectional rnn for medical event detection in electronic health
  records.
\newblock In {\em Proceedings of the conference. Association for Computational
  Linguistics. North American Chapter. Meeting}, volume 2016, page 473. NIH
  Public Access.

\bibitem[Ji and Grishman, 2008]{ji2008refining}
Ji, H. and Grishman, R. (2008).
\newblock Refining event extraction through cross-document inference.
\newblock In {\em Proceedings of ACL-08: Hlt}, pages 254--262.

\bibitem[Krupka et~al., 1991]{krupka1991ge}
Krupka, G., Jacobs, P.~S., Rau, L., and Iwanska, L. (1991).
\newblock Ge: Description of the nltoolset system as used for muc-3.
\newblock In {\em THIRD MESSAGE UNDERSTANDING CONFERENCE (MUC-3): Proceedings
  of a Conference Held in San Diego, California, May 21-23, 1991}.

\bibitem[Lee et~al., 2003]{LEE2003431}
Lee, C.-S., Chen, Y.-J., and Jian, Z.-W. (2003).
\newblock Ontology-based fuzzy event extraction agent for chinese e-news
  summarization.
\newblock {\em Expert Systems with Applications}, 25(3):431 -- 447.

\bibitem[Li et~al., 2013]{li2013joint}
Li, Q., Ji, H., and Huang, L. (2013).
\newblock Joint event extraction via structured prediction with global
  features.
\newblock In {\em Proceedings of the 51st Annual Meeting of the Association for
  Computational Linguistics (Volume 1: Long Papers)}, pages 73--82.

\bibitem[Liao and Grishman, 2010]{liao2010using}
Liao, S. and Grishman, R. (2010).
\newblock Using document level cross-event inference to improve event
  extraction.
\newblock In {\em Proceedings of the 48th Annual Meeting of the Association for
  Computational Linguistics}, pages 789--797. Association for Computational
  Linguistics.

\bibitem[Liu et~al., 2018a]{liu2018event}
Liu, J., Chen, Y., Liu, K., and Zhao, J. (2018a).
\newblock Event detection via gated multilingual attention mechanism.
\newblock In {\em Thirty-Second AAAI Conference on Artificial Intelligence}.

\bibitem[Liu et~al., 2018b]{liu2018jointly}
Liu, X., Luo, Z., and Huang, H. (2018b).
\newblock Jointly multiple events extraction via attention-based graph
  information aggregation.
\newblock {\em arXiv preprint arXiv:1809.09078}.

\bibitem[Maisonnave et~al., 2020]{maisonnave2020assessing}
Maisonnave, M., Delbianco, F., Tohm{\'e}, F., Maguitman, A.~G., and Milios,
  E.~E. (2020).
\newblock Assessing causality structures learned from digital text media.
\newblock In {\em Proceedings of the ACM Symposium on Document Engineering
  2020}, pages 1--4.

\bibitem[Mikolov et~al., 2013a]{mikolov2013efficient}
Mikolov, T., Chen, K., Corrado, G., and Dean, J. (2013a).
\newblock Efficient estimation of word representations in vector space.
\newblock {\em arXiv preprint arXiv:1301.3781}.

\bibitem[Mikolov et~al., 2013b]{mikolov2013distributed}
Mikolov, T., Sutskever, I., Chen, K., Corrado, G.~S., and Dean, J. (2013b).
\newblock Distributed representations of words and phrases and their
  compositionality.
\newblock In {\em Advances in neural information processing systems}, pages
  3111--3119.

\bibitem[Nguyen et~al., 2016a]{nguyen2016joint}
Nguyen, T.~H., Cho, K., and Grishman, R. (2016a).
\newblock Joint event extraction via recurrent neural networks.
\newblock In {\em Proceedings of the 2016 Conference of the North American
  Chapter of the Association for Computational Linguistics: Human Language
  Technologies}, pages 300--309.

\bibitem[Nguyen et~al., 2016b]{nguyen2016two}
Nguyen, T.~H., Fu, L., Cho, K., and Grishman, R. (2016b).
\newblock A two-stage approach for extending event detection to new types via
  neural networks.
\newblock In {\em Proceedings of the 1st Workshop on Representation Learning
  for NLP}, pages 158--165.

\bibitem[Nguyen and Grishman, 2015]{nguyen2015event}
Nguyen, T.~H. and Grishman, R. (2015).
\newblock Event detection and domain adaptation with convolutional neural
  networks.
\newblock In {\em Proceedings of the 53rd Annual Meeting of the Association for
  Computational Linguistics and the 7th International Joint Conference on
  Natural Language Processing (Volume 2: Short Papers)}, pages 365--371.

\bibitem[Nguyen and Grishman, 2016]{nguyen2016modeling}
Nguyen, T.~H. and Grishman, R. (2016).
\newblock Modeling skip-grams for event detection with convolutional neural
  networks.
\newblock In {\em Proceedings of the 2016 conference on empirical methods in
  natural language processing}, pages 886--891.

\bibitem[Nguyen and Grishman, 2018]{nguyen2018graph}
Nguyen, T.~H. and Grishman, R. (2018).
\newblock Graph convolutional networks with argument-aware pooling for event
  detection.
\newblock In {\em Thirty-second AAAI conference on artificial intelligence}.

\bibitem[Patwardhan and Riloff, 2009]{patwardhan2009unified}
Patwardhan, S. and Riloff, E. (2009).
\newblock A unified model of phrasal and sentential evidence for information
  extraction.
\newblock In {\em Proceedings of the 2009 Conference on Empirical Methods in
  Natural Language Processing: Volume 1-Volume 1}, pages 151--160. Association
  for Computational Linguistics.

\bibitem[Peters et~al., 2018]{peters2018deep}
Peters, M.~E., Neumann, M., Iyyer, M., Gardner, M., Clark, C., Lee, K., and
  Zettlemoyer, L. (2018).
\newblock Deep contextualized word representations.
\newblock {\em arXiv preprint arXiv:1802.05365}.

\bibitem[Pustejovsky et~al., 2003]{pustejovsky2003timeml}
Pustejovsky, J., Castano, J.~M., Ingria, R., Sauri, R., Gaizauskas, R.~J.,
  Setzer, A., Katz, G., and Radev, D.~R. (2003).
\newblock Timeml: Robust specification of event and temporal expressions in
  text.
\newblock {\em New directions in question answering}, 3:28--34.

\bibitem[Riloff, 1996a]{riloff1996automatically}
Riloff, E. (1996a).
\newblock Automatically generating extraction patterns from untagged text.
\newblock In {\em Proceedings of the national conference on artificial
  intelligence}, pages 1044--1049.

\bibitem[Riloff, 1996b]{riloff1996empirical}
Riloff, E. (1996b).
\newblock An empirical study of automated dictionary construction for
  information extraction in three domains.
\newblock {\em Artificial intelligence}, 85(1-2):101--134.

\bibitem[Sandhaus, 2008]{sandhaus2008new}
Sandhaus, E. (2008).
\newblock The new york times annotated corpus.
\newblock {\em Linguistic Data Consortium, Philadelphia}, 6(12):e26752.

\bibitem[Sha et~al., 2018]{sha2018jointly}
Sha, L., Qian, F., Chang, B., and Sui, Z. (2018).
\newblock Jointly extracting event triggers and arguments by dependency-bridge
  rnn and tensor-based argument interaction.
\newblock In {\em Thirty-Second AAAI Conference on Artificial Intelligence}.

\bibitem[Surdeanu et~al., 2006]{surdeanu2006hybrid}
Surdeanu, M., Turmo, J., and Ageno, A. (2006).
\newblock A hybrid approach for the acquisition of information extraction
  patterns.
\newblock In {\em Proceedings of the Workshop on Adaptive Text Extraction and
  Mining (ATEM 2006)}.

\bibitem[Tong et~al., 2020]{tong2020image}
Tong, M., Wang, S., Cao, Y., Xu, B., Li, J., Hou, L., and Chua, T.-S. (2020).
\newblock Image enhanced event detection in news articles.
\newblock In {\em AAAI}, pages 9040--9047.

\bibitem[Walker et~al., 2006]{walker2006ace}
Walker, C., Strassel, S., Medero, J., and Maeda, K. (2006).
\newblock Ace 2005 multilingual training corpus.
\newblock {\em Linguistic Data Consortium, Philadelphia}, 57.

\bibitem[Wu et~al., 2014]{wu2014zero}
Wu, S., Bondugula, S., Luisier, F., Zhuang, X., and Natarajan, P. (2014).
\newblock Zero-shot event detection using multi-modal fusion of weakly
  supervised concepts.
\newblock In {\em Proceedings of the IEEE Conference on Computer Vision and
  Pattern Recognition}, pages 2665--2672.

\bibitem[Yangarber et~al., 2000]{yangarber2000automatic}
Yangarber, R., Grishman, R., Tapanainen, P., and Huttunen, S. (2000).
\newblock Automatic acquisition of domain knowledge for information extraction.
\newblock In {\em Proceedings of the 18th conference on Computational
  linguistics-Volume 2}, pages 940--946. Association for Computational
  Linguistics.

\bibitem[Zhang et~al., 2019]{zhang2019joint}
Zhang, T., Ji, H., and Sil, A. (2019).
\newblock Joint entity and event extraction with generative adversarial
  imitation learning.
\newblock {\em Data Intelligence}, 1(2):99--120.

\end{thebibliography}

\newpage
\appendix

\section{Preliminary Experiments for the State-of-the-Art Baseline}

\subsection{Results}

In this section, we describe the preliminary study for the baseline model. Based on this study, we chose the final four variants of the baseline model that were evaluated on the testing data. We conducted this preliminary study for determining the best window size to be used in the baseline model.

For each of the models in the preliminary study, we run five different trials.  We used consecutive seeds from 1 to 5 for each model to guarantee replicability. For each model we report the average sensitivity, specificity, and F1-score of the five trials in both the training and validation data. We do not report accuracy because we are in the presence of highly unbalanced data.  To thoroughly analyze the F1-score on the validation data, we compute two additional metrics. Firstly, we calculate the confidence intervals (CI) at the 95\% level of confidence. Secondly, we compute the p-value for a t-test between each model and the best model reported in each table. We select the best model in terms of the F1-score in the validation data. For example, the p-value of 0.410 in the fourth row of Table \ref{table:CNN1} is indicating that there is no enough evidence to reject the null hypothesis that model 4 has a statistically significantly different F1-score  from the best model of the table (model 1) in the validation data.

We run the baseline model with six different window sizes. Although in the original paper, the authors used a fixed window of size 31, we run experiments to determine if this was the best window size for our setting. We do not use the original window size of 31 because our data was different from the data used in the original paper. While they used full news articles, we only used text fragments (sentences). Therefore, our instances were much smaller. Hence, we required smaller window sizes to avoid using too much padding. Given our setting, a window of size 31 would result in a large amount of padding for many tokens. For example, in a sentence of length 31, only the middle word will not have padding while every other word will have. This extra unnecessary padding will add noise and increase the computational cost of the model with little gain in performance. Guided by this intuition, we conducted preliminary studies to explore different window sizes. 

In table \ref{table:CNN1}, we present the result of the models with the six analyzed window sizes. Since the position embedding is used for representing the relative position inside the window, a window of size 1 does not require a position embeddings. Therefore, we excluded the position embedding from this model. All the other features were used in the reported experiments. 

\subsection{Discussion}

We derive the following conclusions from these preliminary experiments.  As we hypothesized, long window sizes were not suitable for our setting. We find evidence to support this intuition in the results from models 5 and 6. For these models, the specificity reached 1.0, and the F1-score dropped to almost 0. For this reason, we do not select this model for the final experiments on the testing dataset. 

The model with window size 1 (model 1), achieves the best performance with an F1-score of 59.4\% in the validation data. Therefore, we selected this model to use it in the testing data. We also selected model 4 because of its high performance in the validation data. Furthermore, the high p-value provides no evidence to reject the null hypothesis that model 1 has a performance statistically different from this model. Although models 2 and 3 have a decent performance (56.2\% and 55.8\%, respectively), the statistical analysis shows that we can reject the null hypothesis. Therefore, we have evidence to believe that model 1 is statistically better than models 2 and 3.

In summary, the best model of the table is model 1, with a window of size 1. Model 4, with a window of size 11, has comparable performance, and the analysis showed that there is no statistical difference between them. Therefore, we choose these two window sizes for the experiments on the testing data.

\begin{table}
\resizebox{\columnwidth}{!}{%
\begin{tabular}{ccccccccccl}
\hline
\multicolumn{11}{c}{CNNs on the Validation Dataset: Assessing Window Sizes}                                               \\ \hline
\multirow{2}{*}{Model} & \multirow{2}{*}{Win Size} & \multirow{2}{*}{Features} & \multicolumn{3}{c}{train} & \multicolumn{5}{c}{validation} \\ \cmidrule(r){4-6} \cmidrule(l){7-11} 
    &&            & $\overline{sens}$ & $\overline{spec}$ & $\overline{F1}$     & $\overline{sens}$ & $\overline{spec}$ & $\overline{F1}$ & F1 CI  & p-value          \\ \hline

1 &     1           & \{$W$,$E$\}       &   0.608 & 	0.980 & 	0.692 & 	0.502 & 	0.970 & 	\textbf{0.594}  & 	$\pm$ 0.019 & 	---      \\
2 &     3           & \{$W$,$E$,$Po$\}     &   0.749 & 	0.986 & 	0.808 & 	0.466 & 	0.975 & 	0.562           & 	$\pm$ 0.015 & 	0.003    \\
3 &     5           & \{$W$,$E$,$Po$\}     &   0.816 & 	0.988 & 	0.858 & 	0.464 & 	0.975 & 	0.558           &  	$\pm$ 0.023 & 	0.006   \\
4 &     11          & \{$W$,$E$,$Po$\}     &   0.811 & 	0.979 & 	0.852 & 	0.500 & 	0.965 & 	0.590           & 	$\pm$ 0.043 & 	0.410   \\
5 &     21          & \{$W$,$E$,$Po$\}     &   0.015 & 	1.000 & 	0.022 & 	0.002 & 	0.999 & 	0.002           &  	$\pm$ 0.002 & 	0.000\\
6 &     31          & \{$W$,$E$,$Po$\}     &   0.001 & 	1.000 & 	0.001 & 	0.001 & 	1.000 & 	0.001           & 	$\pm$ 0.002 & 	0.000 \\ \hline

\end{tabular}
}
\caption{Average performance of the CNN model for six different window sizes. For each window size, we run five trials and report the average sensitivity, specificity, and F1-score for both the training and validation data. For the validation data, we also report the confidence intervals at a 95\% level of significance for the F1-score. We also report for each model the p-value of a single-tail t-test against the best model of the table. The test is performed between the F1-score of each of the two models in the validation data. A small p-value gives evidence to reject the hypothesis that the models have the same F1-score, indicating that the best model of the table is statistically better.\\
In these experiments, we try six different window sizes to determine the best one for using on the testing dataset. The baseline achieved the best performance with window size 1. We found evidence that suggests that the model is statistically better than the ones with window sizes 3, 5, 21 and 31. Although the model with window size 1 performs slightly better than the one with window size 11, we could not find statistical evidence to suggest that one is better than the other. Therefore, the experiment on the testing dataset were carried out with these two window sizes.} \label{table:CNN1}
\end{table}

\section{Preliminary Experiments for the Proposed \\ Model}

In this section, we describe the preliminary studies carried out to evaluate the proposed model (RNN). Based on these studies, we chose the seven final variant models to use on the testing data. We conducted two preliminary studies. One for determining the number of Bi-LSTM layers and hidden units, and another for studying the features used. 

For each of the models in the preliminary study, we run five different trials. We report the average of the five trials for each model. We used consecutive seeds from 1 to 5 for each model to guarantee replicability. We report for each model, the average sensitivity, specificity, and harmonic mean between these two metrics, namely F1-score, for  both the training and validation data. We exclude the accuracy of the analysis because we are in the presence of highly unbalanced data. To thoroughly analyze the F1-score on the validation data, we compute two additional metrics. Firstly, we calculate the confidence interval (CI) at the 95\% level of confidence for this metric. Secondly, we compute for each model the p-value for a t-test between the model considered and the best model of each table. We define the best model in terms of the F1-score in the validation data. For example, the p-value of 0.160 in the fifth row of Table \ref{table:RNN1} is indicating that there is no enough evidence to reject the null hypothesis that model 5 has a statistically different F1-score from the best model of the table (model 6) in the validation data.

\subsection{Results of the First Preliminary Study for the Proposed Model}
We run the proposed model with eight different architectures (by varying the number of Bi-LSTM layers and hidden units). We configured the number of hidden units to be in descending order, with the first layer having the largest number and the last layer having a smallest number of units. This configuration follows the intuition that each layer should take the input from the previous one and build less but more elaborate features with higher levels of abstraction. For the remaining of this paper, we will describe each architecture as a sorted list of hidden units, where the first number is the number of hidden units of the first BLSM layer, and so on. For example, the architecture $\langle$100,15,5$\rangle$ is a three-layer architecture, with 100 hidden units in the first layer, 15 in the next one, and 5 in the last one. Except for the Bi-LSTM layers, the remaining of the network is the same for the eight models.

In table \ref{table:RNN1}, we present the result of the models for the eight different configurations for the Bi-LSTM layers. The first model has three-layers of Bi-LSTM, with 100, 15, and 5 hidden units in the first, second, and third layers, respectively. The second and third models have two layers of Bi-LTSM each, with varying numbers of hidden units. For the second model, we use 15 and 5 hidden units, while for the third, we use 5 and 2. The remaining five modes are all single-layer, with different numbers of hidden units. We present the models from the most complex to the simplest. Models 4 to 8 have 200, 50, 15, 7, and 1 hidden units, respectively. As we previously mentioned, the remaining of the network is the same for the eight models.

\subsection{Discussion}

%\textbf{We derive the following conclusions from the first preliminary experiments}. Firstly, m
The results achieved by the multiple-layer models 1 and 2 are similar to (69.2\% and 68.2\%, respectively) and considerably worse than those achieved by  model 6 (72.9\%), which is a much simpler model.  Model 3 and 4 are the worst two models, with a performance of 66.8\% and 66.0\%, respectively. These results show the importance of finding the right balance between complexity and simplicity. A three-layer model (model 1) achieves good performance, but not as a good as that achieved by the considerably simpler single-layer models (models 5 to 8). However, too simple models, such as model 3, with a small number of hidden units, also perform poorly. 

The best model is model 6, which is a single-layer model with 15 hidden units, and hence has a good balance between complexity and simplicity. Furthermore, the four best models are all single-layer models, with 50 hidden units or less (50, 15, 7 and 1). From the results achieved by these models, we observe an inverse relation between complexity and performance. Simpler models achieve better performances. These results confirm the intuition that big and complex models require more data to learn general patterns and avoid overfitting. Therefore, in the context of ED, where the datasets are in the order of hundreds of documents, complex models are prone to harm the performance. Guided by this intuition, which is confirmed by the results, we selected the architecture of model 6 for the second preliminary study and the experiments on the testing set.

\begin{table}
\resizebox{\columnwidth}{!}{%
\begin{tabular}{ccccccccccl}
\hline
\multicolumn{11}{c}{RNNs on the Validation Dataset: Assessing Different Architectures}                                               \\ \hline
\multirow{2}{*}{Model} & \multirow{2}{*}{Architecture} & \multirow{2}{*}{Features} & \multicolumn{3}{c}{train} & \multicolumn{5}{c}{validation} \\ \cmidrule(r){4-6} \cmidrule(l){7-11} 
    &&            & $\overline{sens}$ & $\overline{spec}$ & $\overline{F1}$     & $\overline{sens}$ & $\overline{spec}$ & $\overline{F1}$ & F1 CI  & p-value          \\ \hline
    
1& $\langle$100,15,5$\rangle$  & all           &    0.722 & 	0.988 & 	0.742 & 	0.654 & 	0.962 & 	0.692           & 	$\pm$ 0.037 & 	0.026  \\ \relax
2& $\langle$15,5$\rangle$      & all           &    0.732 & 	0.984 & 	0.748 & 	0.648 & 	0.958 & 	0.682           & 	$\pm$ 0.019 & 	0.002  \\ \relax
3& $\langle$5,2$\rangle$       & all           &    0.723 & 	0.981 & 	0.741 & 	0.628 & 	0.964 & 	0.668           & 	$\pm$ 0.053 & 	0.015  \\ \relax  
4& $\langle$200$\rangle$      & all           &    0.741 & 	0.982 & 	0.758 & 	0.630 & 	0.956 & 	0.660           & 	$\pm$ 0.043 & 	0.004  \\ \relax
5& $\langle$50$\rangle$       & all           &    0.729 & 	0.982 & 	0.747 & 	0.685 & 	0.950 & 	0.704           & 	$\pm$ 0.059 &   0.160  \\ \relax
6& $\langle$15$\rangle$       & all           &    0.757 & 	0.977 & 	0.765 & 	0.709 & 	0.952 & 	\textbf{0.729}  & 	$\pm$ 0.026 & 	---    \\ \relax
7& $\langle$7$\rangle$        & all           &    0.749 & 	0.983 & 	0.763 & 	0.681 & 	0.948 & 	0.698           & 	$\pm$ 0.039 & 	0.052  \\ \relax
8& $\langle$1$\rangle$        & all           &    0.763 & 	0.977 & 	0.771 & 	0.694 & 	0.945 & 	0.708           & 	$\pm$ 0.026 & 	0.070  \\ \hline

\end{tabular}
}
\caption{Average performance of the proposed model (RNN) for eight different architectures. The architecture is depicted with a list that represents the number of hidden units of each layer. The architecture $\langle$15,5$\rangle$ is a network with two Bi-LSTM layers with 15 and 5 hidden units in the first and second layers, respectively. The layers before and after the Bi-LSTM layers do not vary. 
For each architecture, we run five trials and report the average sensitivity, specificity, and the harmonic mean between these two metrics, namely F1-score, on both the training and validation data. For the validation data, we also report the confidence interval at a 95\% level of significance for the F1-score. We also report the p-value of a single-tail t-test of each model against the best model of the table. The test is performed between the F1-score for each of the two models on the validation data. A small p-value gives evidence to reject the hypothesis that the models have the same F1-score, indicating that the best model of the table is statistically better.\\
In these experiments, we tried eight different number of architectures (number of Bi-LSMT layers and hidden units) for our proposed model. Model 6, the single-layer model with 15 hidden units, is the one with the best performance. The four models with the best results are single-layer, indicating that in this context of small datasets, simpler models perform better. Models that are too complex, such as model 4, with 200 hidden units, and models that are too simple, such as model 3, with only two small Bi-LSTM layers, perform poorly. The results provide evidence to believe model 6 achieves the best balance between complexity and simplicity. Therefore, we use the architecture of model 6 for the subsequent experiments.}  \label{table:RNN1}
\end{table}

\subsection{Results of the Second Preliminary Study for the Proposed Model}
We performed a second preliminary study by varying the features used in the model to study the impact of each feature on the overall performance.  We conducted these experiments in a similar way to an ablation study. We first measure the performance of the model with all the features, and for each model or hypothesis we wanted to test, we run a new model with different features removed and compared its performance with that of the model that maintained all the features. 

The goal of this preliminary experiment was to assess the impact of each feature on the overall performance. This allows us to remove those features that are not useful for the task or that add noise and therefore harm the performance. We performed five trials for each set of features using the best architecture found in the previous preliminary study. The results of these experiments are in Table \ref{table:RNN2},  where the first row reports the performance of the model with all the features, and the following are the same model with one or more features removed. 

We tested nine different feature configurations, including  model 1, which contained all the features (all). To assess the impact of contextual embeddings we evaluated three different models (models 2 to 4). Model 2 contains all the features except for the contextual embeddings (all-\{$B$\}). Model 3 includes all but the contextual sentence embeddings (all-\{$S$\}), which are features constructed by adding up the contextual embeddings ($B$) for all the words in the sentence. Finally, model 4 has both the contextual word and contextual sentence embeddings removed (all-\{$B$,$S$\}).

We also evaluated two different models to measure the impact of the use of grammatical information. First, we evaluated model 5, where the Part-Of-Speech tags are removed, both the simplified version and the detailed ones (all-\{$P$,$T$\}). Second, we evaluated model 6, where the dependency tags were removed (all-\{$T$\}).

Finally, we evaluated three additional models to test the impact of the three remaining features: the entity embeddings ($E$), the Word2Vec embeddings ($W$), and Spacy contextual word embeddings ($Sp$). These are models 7, 8, and 9, respectively.

\begin{table}
\resizebox{\columnwidth}{!}{%
\begin{tabular}{ccccccccccl}
\hline
\multicolumn{11}{c}{RNNs on the Validation Dataset: Assessing Feature Contribution}                                               \\ \hline
\multirow{2}{*}{Models} & \multirow{2}{*}{Architecture} & \multirow{2}{*}{Features} & \multicolumn{3}{c}{train} & \multicolumn{5}{c}{validation} \\ \cmidrule(r){4-6} \cmidrule(l){7-11} 
    &&            & $\overline{sens}$ & $\overline{spec}$ & $\overline{F1}$     & $\overline{sens}$ & $\overline{spec}$ & $\overline{F1}$ & F1 CI  & p-value          \\ \hline
1& $\langle$15$\rangle$    &       all         &       0.757 & 	0.977 & 	0.765 & 	0.709 & 	0.952 & 	0.729            & 	$\pm$ 0.026 & 	0.326       \\ \relax
2& $\langle$15$\rangle$    &       all-\{$B$\}   &       0.707 & 	0.980 & 	0.728 & 	0.528 & 	0.963 & 	0.578            & 	$\pm$ 0.072 & 	0.001     \\ \relax
3& $\langle$15$\rangle$    &       all-\{$S$\}   &       0.671 & 	0.995 & 	0.710 & 	0.571 & 	0.971 & 	0.626            & 	$\pm$ 0.061 & 	0.003     \\ \relax
4& $\langle$15$\rangle$    &       all-\{$B$,$S$\} &       0.781 & 	0.997 & 	0.790 & 	0.594 & 	0.935 & 	0.637            & 	$\pm$ 0.023 & 	0.000     \\ \relax
5& $\langle$15$\rangle$    &       all-\{$T$,$P$\} &       0.752 & 	0.976 & 	0.761 & 	0.667 & 	0.957 & 	0.698            & 	$\pm$ 0.037 & 	0.026       \\ \relax
6& $\langle$15$\rangle$    &       all-\{$D$\}   &       0.747 & 	0.980 & 	0.758 & 	0.718 & 	0.944 & 	0.733            & 	$\pm$ 0.017 & 	0.430       \\ \relax
7& $\langle$15$\rangle$    &       all-\{$E$\}   &       0.761 & 	0.977 & 	0.769 & 	0.697 & 	0.950 & 	0.721            & 	$\pm$ 0.034 & 	0.182     \\ \relax
8& $\langle$15$\rangle$    &       all-\{$W$\}   &       0.629 & 	0.973 & 	0.644 & 	0.707 & 	0.954 & 	0.727            & 	$\pm$ 0.049 & 	0.341      \\ \relax
9& $\langle$15$\rangle$    &       all-\{$Sp$\}  &       0.748 & 	0.979 & 	0.759 & 	0.721 & 	0.946 & 	\textbf{0.735}   & 	$\pm$ 0.018 & 	---       \\ \hline

\end{tabular}
}
\caption{Average performance of the proposed model (RNN) for nine different sets of features using the best architecture (single Bi-LSTM layer with fifteen hidden units). For each set of features, we run five trials and report the average sensitivity, specificity, and F1-score in both the training and validation data. For the validation data, we also report the confidence intervals at a 95\% level of significance for the F1-score. We also report for each model the p-value of a single-tail t-test against the best model of the table. The test is performed between the F1-score in the validation data for each of the two models. A small p-value gives evidence to reject the hypothesis that the models have the same F1-score, indicating that the best model of the table is statistically significantly better.\\
In these experiments, we examine nine different sets of features. We use the complete set of features for comparison, and remove each different input to measure its impact. We removed T and P together because they are semantically the same input (i.e., the simplified and the detailed versions of Part-Of-Speech). And we remove $B$ and $S$ inputs together because those are the two contextual-embedding-related inputs. The contextual-embedding-related features show the most significant impact. The results show that features $D$, $E$, $W$, and $Sp$ have a negligible impact on the performance. We can conclude from these results that, in the presence of all the other features, we can remove each of these features without harming the performance. On the other hand, the T and P features have a small but statistically significant effect when removed. We conducted further experiments to find additional evidence for these findings.}  \label{table:RNN2}

\end{table}

\subsection{Discussion}

We observe that the performance drops considerably (a drop of 15.1\% in F1-score) for the model that does not contain the contextual word embeddings (model 2) in comparison to the model that contains all the features (model 1). Similarly, the model without the contextual sentence embeddings (model 3) has a drop in performance of 10.3\%. On the other hand, the performance of the model without both features (model 4) has a performance similar to that of the model where only the contextual sentence embeddings are removed (model 3). A statistical analysis shows that for the three models, the hypothesis that they are equal to the best model can be rejected with a confidence level of over 95\%. The two individual analyses of the contextual embeddings show the significant impact of each in the overall performance and indicate a slightly higher impact of the contextual word embeddings over the contextual sentence embeddings.

The results for the model without the dependency information ($D$) (model 6) are similar to those obtained by the model that preserves all the features (model 1). These results provide evidence that suggests that the model could not take advantage of the information of this input. Three factors can be influencing these results. By removing the Part-Of-Speech information ($T$, $P$) (model 5), we have a small drop in performance (3.1\%). This result suggests that the feature can be useful for the ED task. We further explore the inclusion and exclusion of all this grammatical information ($T$, $P$, $D$) in the testing set.

%discussion
Rows 7 to 9 of table \ref{table:RNN2}, corresponding to models 7 to 9, show the models with all the features excluding the entity embeddings ($E$), the Word2vec word embedding ($W$), and the Spacy contextual embeddings ($Sp$), respectively. Model 9 is the one with the best performance. However, the high p-values show that there is no statistical difference between the best model and the other two. Furthermore, there is no statistical difference between the best model and the model with all the features (model 1). These results show that removing any of these three features ($E$, $W$, $Sp$) has a negligible impact on the overall performance. We further study the effect of including and excluding these features in the testing set. However, based on these results, it is expected that, in the presence of the other attributes, excluding these three features improves performance.

\end{document}